\documentclass[runningheads]{llncs}
\pdfoutput=1 
\usepackage{graphicx}
\usepackage{multirow}
\usepackage{multicol}
\usepackage[dvipsnames]{xcolor}
\begin{document}
\title{Utilizing Natural Language Processing for Automated Assessment of Classroom Discussion\thanks{Supported by a grant from the Learning Engineering Tools Competition.}}
\titlerunning{Automated Assessment of Classroom Discussion}
%
\author{Nhat Tran \and Benjamin Pierce \and Diane Litman \and Richard Correnti \and Lindsay Clare Matsumura}
\authorrunning{Tran et al.}
%
\institute{University of Pittsburgh, Pittsburgh, USA \\
\email{\{nlt26,bep51,dlitman,rcorrent,lclare\}@pitt.edu}\\
}

%
\maketitle              
\begin{abstract}
Rigorous and interactive class discussions that support students to engage in high-level thinking and reasoning are essential to learning 
and are a central component of most teaching interventions. However, formally assessing discussion quality ‘at scale’ is expensive and infeasible for most researchers. In this work, we experimented with various modern natural language processing (NLP)  techniques to automatically generate rubric scores for individual dimensions of classroom text discussion quality. Specifically, we worked on a  dataset of 90 classroom discussion transcripts consisting of over 18000 turns annotated with fine-grained Analyzing Teaching Moves (ATM) codes and focused on four Instructional Quality Assessment (IQA) rubrics. 
Despite the limited amount of data, our work shows encouraging results in some of the rubrics while suggesting that there is room for improvement in the others. We also found that certain NLP approaches work better for certain rubrics.

\end{abstract}
\section{Introduction and Background}
\label{sec:background}
Instructional quality has been of great interest to educational researchers for 
several decades. Due to their cost, measures of instructional quality that can be obtained at-scale remain elusive. 
Previous work \cite{talkmoves} has shown that providing automated feedback on teachers' talk moves can lead to positive instructional changes.
We report here on initial attempts 
to apply Natural Language Processing (NLP) methods such as pre-trained language models or sequence labeling with BiLSTM \cite{lstm_crf_2} to automatically produce rubric scores for individual dimensions of classroom discussion quality from transcripts,
\begin{table} [t!]
\centering
\caption{Data distribution and mean (\textbf{Avg}) of 4 focused {\it IQA} rubrics for Teacher ($T$) and Student ($S$) with their relevant \textit{ATM} codes. An {\it IQA} rubric's distribution is represented as the counts of each score (1 to 4 from left to right) (n=90 discussions).}
\label{tab:data}
\begin{tabular}{l|c|c|c|c}
\multicolumn{3}{c|}{\textbf{Rubric}} & \multicolumn{2}{c}{\textbf{Relevant \textit{ATM} code}}\\
\hline
\hfil \textbf{Short Description} & \textbf{Distribution} & \textbf{Avg} & \textbf{Code Label} & \textbf{Count} \\
\hline
S1: \textit{T} connects \textit{S}s & [51, 19, 9, 11] & 1.8 & Recap or Synthesize S Ideas&  75\\
\hline
S2: \textit{T} presses $S$& [7, 7, 9, 67] & 3.6& Press & 927 \\
\hline
S3: $S$ builds on other's idea& [65, 6, 8, 11]& 1.6 & Strong Link& 101 \\
\hline
\multirow{2}{*}{S4: $S$ support their claims}& \multirow{2}{*} {[28, 12, 8, 42]} & \multirow{2}{*}{2.7}& Strong Text-based Evidence & 403 \\
\cline{4-5}
& & & Strong Explanation & 286 \\
\hline
 \multicolumn{2}{c|} - & & Others & 52687 \\
\end{tabular}
\end{table}
building upon
two established measures that have shown high levels of reliability and validity in prior learning research – the {\it Instructional Quality Assessment (IQA)} and the {\it Analyzing Teaching Moves (ATM)} rubrics 
\cite{correnti_2,lindsay_1}.

Our corpus consists of 170 videos
from 31 
English Language Arts classrooms in a Texas district. 
18 teachers taught fourth grade,  13 taught fifth grade, and on average had 13 years of teaching experience. 
The 
student population 
was considered low income (61\%), with 
students identifying as: Latinx (73\%), Caucasian (15\%), African American (7\%), multiracial (4\%), and Asian or Pacific Islander (1\%).
The 
videos were  manually scored holistically, on a scale from 1 to 4, using the \textit{IQA}  on 11 
dimensions 
for both teacher and student contributions. They were also scored using more fine-grained talk moves at the sentence level using the \textit{ATM} 
discourse measure.
The current work 
focuses on only the \textbf{90} discussion transcripts that have already been converted to text-based codes. 

As summarized in column 1 of Table~\ref{tab:data}, 
our classifiers are trained  to predict 4 of the 11 \textit{IQA} rubrics containing aspirational teacher (T) and student (S) ‘talk moves’ – {\it T Links Student Contributions} (score S1), {\it T Presses for Information} (S2),  {\it S Link Contributions} (S3), {\it S Support Claims with Evidence} (S4). 
Besides the 5 \textit{ATM} codes (column 4 in Table \ref{tab:data}) related to these 4 \textit{IQA}  rubrics 
the rest are labeled as \textit{Others}. The distributions of \textit{IQA}  scores for each rubric and of relevant \textit{ATM} codes are summarized in columns 2 and 5 of Table \ref{tab:data}, respectively. We notice that the frequencies of ATM codes related to S1 and S3 are very low (less than or approximately 1 per transcript).
Below is an example excerpt with annotated \textit{ATM} codes from our data:

\begin{quote}
    \textbf{Teacher}: [The girls get the water and the boys do the herds, right?]\textsubscript{\textcolor{teal}{Others}}
    [Where did you get that from the text?]\textsubscript{\textcolor{blue}{Press}}
     
     \textbf{Student}: [Other people, mostly women and girls who had to come fill their own containers, many kinds of birds, all flap, twittering and cawing. Herds of cattle had been brought to good grazing by the young boys who looked after them.]\textsubscript{\textcolor{red}{Strong Text-Based Evidence}}
\end{quote}

In this paper we present several  {\it IQA} classifiers, and show that using predicted {\it ATM} codes as features outperforms an end-to-end model. The long-term goal of our work is to use such classifiers in  a tool for automated {\it IQA} assessment  so that teachers and coaches can evaluate classroom discussion quality in real-time.

\section{Methods}
We  train different 
models for \textit{IQA}  assessment to explore tradeoffs between scoring performance,  explainability, and  training set.  Our baseline is a neural {\bf  end-to-end 
model}, as neural models
often have high 
performance and do not require  feature engineering. However, since
the \textit{IQA}  score of a specific rubric can be inferred from the number of times the relevant \textit{ATM} codes are used (Section \ref{sec:background}),
we also develop \textbf{\textit{IQA} prediction models using \textit{ATM} codes} as predictive features.  This in turn requires 
 \textit{ATM} models to predict the 
 relevant 6 \textit{ATM} codes.  Specifically, for each sentence, these  models will predict 1 of 6 \textit{ATM} code labels in column 4 of Table \ref{tab:data}.
 This is a 6-way classification task.

\textbf{Hierarchical \textit{ATM} Classification.}
We hypothesize that it would be easier to separate \textit{Others} from the 5 focal \textit{ATM} codes as they have specific usages.
We perform a 2-step hierarchical classification at sentence level as follows.
Step 1, binary classification, is to classify \textit{Other} versus \textit{5 focal ATM Codes}. 
If the code is not \textit{Others}, step 2 is to perform another 5-way classification to identify the final label. 
We train separate BERT-based classifiers for each step. The input for the classifiers is the current sentence concatenated with previous sentences in the same turn and sentences from one previous turn. Because each \textit{ATM} code except \textit{Others} can be only from one speaker, either Teacher or Student, we train two classifiers for the 5-way classification  of Step 2, one classifier used to predict teacher codes (\textit{Recap or Synthesize S Ideas} and \textit{Press}) and one classifier specialized in student codes (\textit{Strong Link}, \textit{Strong Text-Based Evidence} and \textit{Strong Explanation}). Depending on  the speaker, only one of them is called for Step 2.

\textbf{\textit{ATM} Sequence Labeling.}
A classroom discussion can be considered as a sequence of sentences. This approach assigns a label (1 out of the 6 \textit{ATM} codes) to each sentence in a conversation. Unlike the Hierarchical Classification approach that predicts the label of each sentence independently, in this approach, the label of a given sentence is more dependent on the labels of nearby sentences. We use BERT-BiLSTM-CRF as our sequence labeling model. BiLSTM-CRF has been widely used for sequence labeling tasks \cite{lstm_crf_2} 
and BERT \cite{bert} provides a powerful tool for sentence representation that can work well with that architecture.

\textbf{Additional Techniques.}
During \textit{ATM} classification, since \textit{Others} constitutes more than 90\% of the total labels, we {\bf downsample} the training data to reduce the imbalance.
For \textit{IQA}  classification, annotators tend to group consecutive sentences sharing the same functionality in one turn as one \textit{ATM} code (e.g., one \textit{Strong Texted-Based Evidence} code is used for two sentences in the excerpt in Section \ref{sec:background}), but our prediction is done on sentence level. 
{\bf Merging adjacent \textit{ATM} predictions that are the same into one code} in the inference phase helps preserve this nature.
Also, since the range of the \textit{IQA}  scores is very small (1 to 4), translating the absolute counting of \textit{ATM} codes to \textit{IQA} scores (see Section \ref{sec:background}) can drastically shift the \textit{IQA} scores due to misclassification of the \textit{ATM} codes.
  To alleviate this sensitivity, we  build separate {\bf linear regression models to estimate each \textit{IQA}  score from the counting of relevant \textit{ATM} codes}, then use the nearest integers as the \textit{IQA}  scores.
 \begin{table} [t]
\centering
\caption{\textit{ATM} Codes 6-way Classification Results ($\mathrm{F_1}$ scores over 5-fold cross-validation with standard deviations in parentheses).}
\label{tab:atm}
\begin{tabular}{l|c|c}
\hfil \textbf{Method} & \textbf{Step 1} & \textbf{6-way}\\
\hline
Non-hierarchical Classification (All Data) & - & 0.29 (0.04)\\
\hspace{4mm} \textit{/w} 60\% downsampling & - & 0.49 (0.03) \\
\hline
Hierarchical Classification (All Data) & 0.56 & 0.41 (0.04)\\
\hspace{4mm} \textit{/w} 60\% downsampling & 0.72 & 0.65 (0.02)\\
\hline
Sequence Labeling (All Data)& - & 0.45 (0.03)\\
\hspace{4mm} \textit{/w} 60\% downsampling & - & 0.62 (0.01)\\
\hline
\hline
Hierarchical with Oracle for Step 1 (All Data) & 1 & 0.57 (0.02)\\
\hspace{4mm} \textit{/w} 60\% downsampling & 1 & 0.68 (0.04) \\
\end{tabular}
\end{table}
\section{Results}

\textbf{\textit{ATM} Code Prediction} results (macro average $\mathrm{F_1}$ scores) 
are shown in Table \ref{tab:atm}.
The Non-hierarchical baseline is a BERT-based 6-way classifier given the same input as our hierarchical approach.
Both Hierarchical Classification 
and 
Sequence Labeling 
outperform the  Non-hierarchical baseline, 
 whether using all data or downsampled data for training. 
The numbers also show that downsampling the proportion of the most popular class (\textit{Others}) to 60\% increases the performance of all models\footnote{We tried different ratios and 60\% provides the best results.}. 
For the 2-step Hierarchical Classification approach, it improves the performances of both step 1 
and the final 6-way classification. 
Additionally, the Sequence Labeling  and Hierarchical approaches perform similarly. Although Sequence Labeling has a slightly higher score when all training data is used (0.45 vs. 0.41, $\rho = 0.039$), it is inferior to Hierarchical Classification with 60\% downsampling  (0.62 vs. 0.65, $\rho = 0.046$).
Using a perfect Oracle model with 100\% accuracy for Step 1 (\textit{Others} versus 
 \textit {5 ATM codes}) does not lead to a large gain in the 6-way classification results compared to
 our fully automated Hierarchical approach in the downsampled version (0.68 vs 0.65).
We thus
use the non-oracle models built with 60\% downsampling rate for the inference of the classroom discussion quality (\textit{IQA}  scores) below.

\textbf{\textit{IQA} Score Prediction} is performed
based on the models for \textit{ATM} codes prediction. The Quadratic Kappa (QK) scores for the estimations of the four rubrics of classroom discussion quality are reported in Table \ref{tab:iqa}. The baseline for each rubric is an end-to-end Longformer model \cite{Longformer} which directly predicts the \textit{IQA}  scores given the raw text transcripts using a linear layer on top of the hidden representation of [CLS], ignoring the \textit{ATM} codes. 
The results show that all 
variations of Hierarchical Classification and Sequence Labeling 
outperform the baselines, which emphasizes the importance of utilizing  \textit{ATM} codes to infer  \textit{IQA}  scores.  Besides increasing performance, the ATM-based models also increase model interpretability, useful for generating formative feedback in the future.\begin{table} [t]
\centering
\caption{\textit{IQA}  Scores Estimation Results in Quadratic Kappa (QK) averaged over 5-fold cross-validation, inferred from Absolute Counting (A.Count) and Linear Regression (Regression) after \textit{ATM} prediction. \textbf{Bold} numbers are the best results for each rubric.}
\label{tab:iqa}
\begin{tabular}{|l|c|c|c|c|c|}
\hline
\hfil \multirow{2}{*}{\textbf{Rubric}} & \multirow{2}{*}{\textbf{Baseline}} & \multicolumn{2}{c|}{\textbf{Hierarchical}} & \multicolumn{2}{c|}{\textbf{Sequence}} \\
\cline{3-6}
& & {A.Count} & {Regression} & {A.Count} & {Regression}\\
\hline
S1: Teacher connects Students & \multirow{2}{*}{0.34} & 0.43 & 0.54 & 0.50 & 0.55 \\
\hspace{4mm} \textit{/w merged codes} & & 0.48 & 0.55 & 0.52 & \textbf{0.57}  \textbf{} \\
\hline
S2: Teacher presses Student & \multirow{2}{*}{0.35} & 0.60  & 0.65 & 0.55 & 0.62\\
\hspace{4mm} \textit{/w merged codes} &  & 0.64 & \textbf{0.68} & 0.57 & 0.63 \\
\hline
S3: Student builds on each other & \multirow{2}{*}{0.30} & 0.42 & 0.51 & 0.47 & 0.51 \\
\hspace{4mm} \textit{/w merged codes} &  & 0.49 & \textbf{0.54} & 0.50 & 0.53 \\
\hline
S4: Student support their claims & \multirow{2}{*}{0.36}  & 0.60 & 0.65 & 0.57 & 0.61 \\
\hspace{4mm} \textit{/w merged codes} &  & 0.65 & \textbf{0.70} & 0.61 & 0.63 \\
\hline
\end{tabular}
\end{table}


One notable observation is that using regression to estimate the \textit{IQA}  scores is always 
better than using absolute counting.  
 This supports our assumption that regression will alleviate the sensitivity of miscounting and provides a smoother transition from the number of times \textit{ATM} codes appear to the actual \textit{IQA}  scores. Using regression, the highest gain in QK scores for Hierarchical Classification (0.09) and Sequence Labeling (0.07) are from S3 and S2, respectively.

 Merging consecutive same \textit{ATM} codes into one also improves the performance of \textit{IQA}  score estimation as expected. For the same approach, the increases from this technique are mostly larger when absolute counting is used. 
 The \textit{ATM} classification was on sentence level and absolute counting is more sensitive to over-counting, so the merging technique is more effective for this method.

 While Sequence Labeling yields the best S1 results, the best results for the other 
 rubrics come from Hierarchical Classification. Our reasoning is that for S1, 
the relation to adjacent sentences plays a more important role to identify the relevant \textit{ATM} code 
 as there should be multiple students speaking out their ideas before the teacher can connect/synthesize them. Thus, Sequence Labeling, which focuses more on  dependencies between sentences, performs better.
 This suggests that certain approaches are more suitable for certain rubrics.
\begin{figure*}[t!]
\includegraphics[width=12cm]{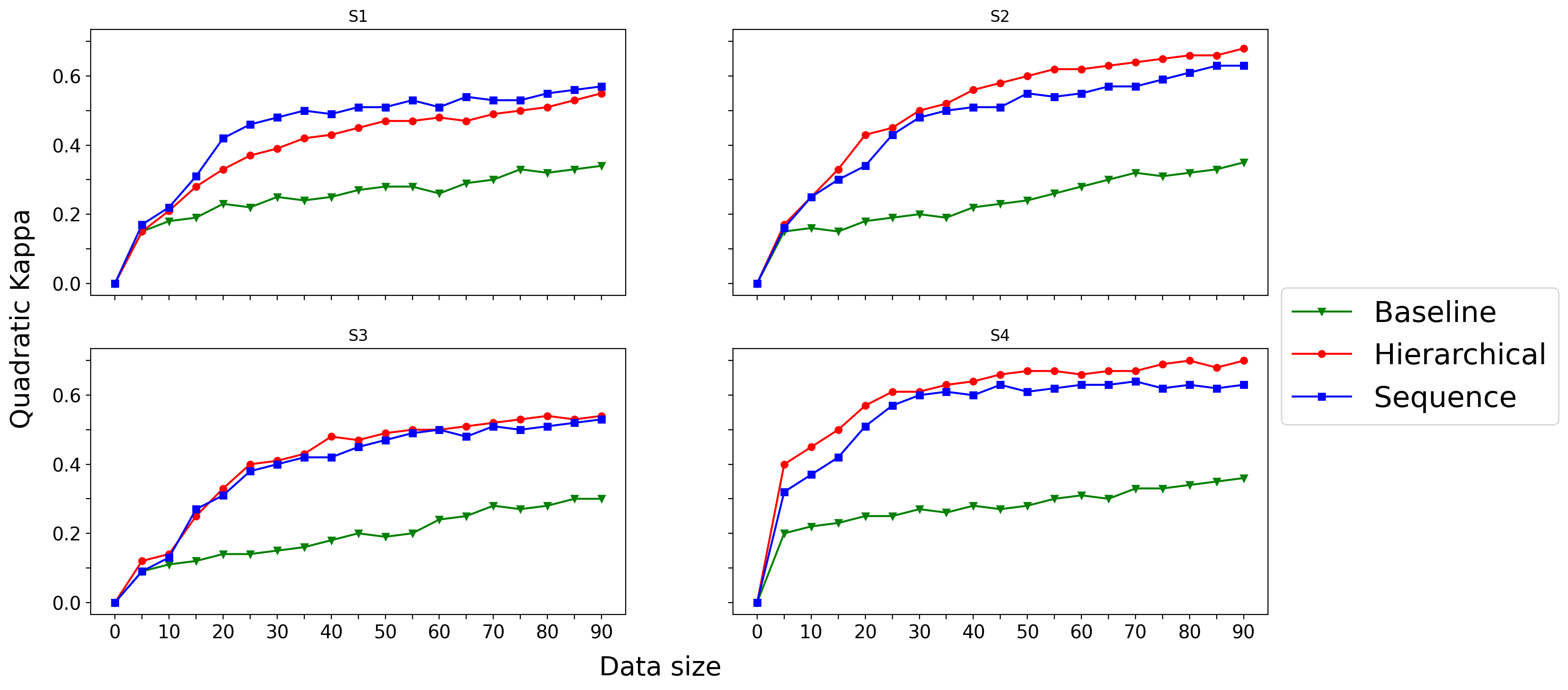}
\caption{\textit{IQA}  Score Estimation Results (QK) in relation to the amount of training data.}
\label{fig:learning_curve}
\end{figure*}

Finally, Figure \ref{fig:learning_curve} demonstrates the performance of our best models (with \textit{regression} and \textit{merged codes}) over training size. 
While the baselines do not improve much  after a certain size of training data, 
the lines of Hierarchical Classification and Sequence Labeling maintain upward trends, 
suggesting that these models will continue to benefit from more data as we complete our video transcription.  Even using only 90 discussions, the QK results show that \textit{ATM}-based model reliability  is already substantial for S2 and S4, and moderate for S1 and S3, even though there were infrequent instances of relevant codes in the corpus.

\section{Conclusion and Future Directions}
We experimented with NLP approaches to automatically 
assess discussion quality using the \textit{IQA}, and
to deal with imbalanced \textit{ATM} data and imperfect \textit{ATM} code prediction.
Our results show that \textit{IQA}  models using either Hierarchical Classification or Sequence Labeling to first predict \textit{ATM} codes outperform baseline end-to-end \textit{IQA}  models, while each of the ATM-based \textit{IQA}  models performs better than the other in certain \textit{IQA}  rubrics. 
Once the full corpus is available, we will generate a validity argument 
for whether automated scoring replicates 
known associations in the corpus, and incorporate the demographic information into our analyses. We will  also utilize \textit{ATM} codes beyond the 5 relevant to the focused rubrics to add more context  for \textit{ATM} prediction. To mitigate the limited size of even the full corpus,  we will explore whether  techniques such as 
transfer learning that can take advantage of classroom discussion data from math
\cite{suresh2021using} or high school \cite{olshefski2020discussion} that are now being made available to the community. 


%
%
%
\bibliographystyle{splncs04}
\bibliography{mybib}
%




\end{document}